\documentclass[11pt]{article}

\usepackage[preprint]{acl}
\usepackage{times}
\usepackage{latexsym}
\usepackage[T1]{fontenc}
\usepackage[utf8]{inputenc}
\usepackage{microtype}
\usepackage{booktabs}
\usepackage{tabularx}
\usepackage{array}

\newcolumntype{Y}{>{\raggedright\arraybackslash}X}

\title{When LLM Essays Outscore Student Essays:
What a Korean Writing Rubric Rewards and Where Readers Disagree}

\author{
\quad Shinwoo Park  \quad Yo-Sub Han$^{\dagger}$ \\
\vspace{0.3cm}
Yonsei University, Seoul, Republic of Korea \\
\texttt{\small pshkhh@yonsei.ac.kr, emmous@yonsei.ac.kr}
}

\newcommand{\correspondingfootnote}{
    \let\oldthefootnote=\thefootnote
    \renewcommand{\thefootnote}{}
    \footnotetext{$\dagger$ Corresponding author.}
    \let\thefootnote=\oldthefootnote
}

\begin{document}
\maketitle

\correspondingfootnote 

\begin{abstract}
LLMs now help students plan, draft, and revise essays. Educational assessment
therefore faces a basic question: how should student and LLM writing be
compared? Rubrics assign points to content, organization, and expression.
Their total can still hide which criteria drive the comparison, where ratings
approach the maximum, and where readers disagree. We therefore conducted a
secondary, post hoc audit of
a Korean writing study with source-informed scoring. Three Korean language and
literature majors used a 16-criterion, 100-point rubric to score six student
essays and 24 essays from four LLMs prompted for three school levels. After
source disclosure, they helped develop the rubric and, according to the
protocol, scored the Phase~2 essays without gold labels while recording source
judgments. The pooled LLM
mean exceeds the student mean by 18.35 points. Orthographic norms and
genre-appropriate register contribute 8.19 points, or 44.7\% of the gap. All
72 LLM ratings reach the orthographic maximum, and 70 reach the register
maximum. The student source mean is highest or tied highest on both creativity
criteria and natural Korean phrasing, where reader agreement is weak.
Criterion-level auditing therefore offers a deeper account of human and LLM
writing than the total alone.
\end{abstract}

\section{Introduction}

Students now have access to LLMs that can generate and revise complete essays,
raising questions about authorship, writing quality, and learning. Studies
compare student and LLM essays, test whether teachers and students can identify
their source, and examine how source information affects ratings
~\citep{herbold-etal-2023-large,waltzer-etal-2023-testing,
fleckenstein-etal-2024-teachers}.

Rubrics make evaluation criteria explicit. A composite score still combines
rubric priorities, weights, scales, and reader judgments. A higher LLM total
can therefore be mistaken for uniformly stronger writing when it reflects a
narrower set of formal criteria. Writing-assessment scores depend on the
construct, population, and intended use~\citep{kane-2013-validating}.
Generated-language ratings depend on readers and study design
~\citep{van-der-lee-etal-2019-best,clark-etal-2021-thats,
elangovan-etal-2024-considers}.

\begin{table}[t]
\centering
\small
\setlength{\tabcolsep}{5pt}
\renewcommand{\arraystretch}{1.08}
\begin{tabularx}{\columnwidth}{@{}Y Y@{}}
\toprule
\multicolumn{1}{c}{\shortstack{\textbf{Higher pooled}\\\textbf{LLM scores}}} &
\multicolumn{1}{c}{\shortstack{\textbf{Highest or tied}\\\textbf{student mean}}} \\
\midrule
\textbf{Formal features and organization} &
\textbf{Creativity and natural Korean phrasing} \\
Orthography, register, and organization &
Both creativity criteria and natural Korean phrasing \\
\addlinespace[2pt]
Formal ratings cluster near the maximum &
Reader agreement is weak on these three criteria \\
\midrule
\multicolumn{2}{@{}>{\centering\arraybackslash}p{\columnwidth}@{}}{
\textbf{Takeaway:} the higher total reflects what this rubric rewards, not
stronger writing in every respect.} \\
\bottomrule
\end{tabularx}
\caption{Three Korean language and literature majors evaluated all 30 Phase~2
essays using the same 100-point rubric
(Section~\ref{sec:protocol-rubric}). LLM essays receive more points for formal
features and organization on average, while student essays have the highest or
tied-highest source mean for both creativity criteria and natural Korean phrasing, where
reader agreement is weak.}
\label{tab:paper-at-a-glance}
\end{table}

Sociodemographic persona prompts produce measurable style
differences~\citep{malik-etal-2024-empirical}, while student simulation separates
cognitive ability from writing proficiency~\citep{chen-etal-2026-cpt}.
Neither shows that a model writes at its assigned school level or explains how
rubric design and reader disagreement shape the score.

We address this gap through a secondary audit of a Korean reader study using
essays from the KatFish dataset released with
KatFishNet~\citep{park-etal-2025-katfishnet}. Three undergraduates majoring in
Korean language and literature used a study-specific 100-point rubric to
evaluate six student essays and 24 LLM essays from four models prompted for
three school levels. Within this fixed sample and source-informed scoring
protocol, we ask two questions:
(RQ1) How consistent is the student--LLM total-score difference across models,
readers, and school levels?
(RQ2) Which criteria produce the total-score gap, where do
ratings approach ceiling, and where do readers disagree?

Table~\ref{tab:paper-at-a-glance} summarizes the central result. We decompose
the 18.35-point gap across all 16 criteria and examine rating concentration,
absolute agreement, source-mean rankings by reader, and school-level profiles.
The analysis explains how this rubric produces the pooled LLM total without
treating it as evidence that the LLM essays are stronger in every respect or
match their assigned school levels. It yields one reporting principle: identify which criteria create
the total, where ratings approach ceiling, and where readers disagree.
Full scoring anchors, criterion means, and reader-level patterns
appear in Appendices~\ref{app:rubric}, \ref{app:criterion-means},
and~\ref{app:reader-heterogeneity}.

\section{Related Work}
\label{sec:related_work}

\paragraph{Linguistic profiling of generated text.}
Linguistic profiling characterizes human-written and LLM-generated texts
beyond a binary source decision. An English benchmark spanning eight domains
and 11 LLMs finds simpler syntax, more diverse semantic content, and greater
feature variation in human-written text
~\citep{zanotto-aroyehun-2025-linguistic}. Korean-focused work
identifies source-associated word-spacing, part-of-speech, and punctuation
patterns~\citep{park-etal-2025-katfishnet}. These studies establish measurable
textual contrasts. The open question is how readers value those contrasts when
a scoring rubric converts them into points.

\paragraph{Human expertise and conditioned evaluation.}
Detection performance differs across reader groups and task designs.
Annotators who frequently use LLMs for writing can make accurate source
judgments and provide detailed rationales
~\citep{russell-etal-2025-people}, whereas boundary detection remains difficult
and varies across readers~\citep{dugan2023real}. In self-presentation tasks,
readers also share flawed source heuristics~\citep{jakesch-etal-2023-human}.
Multilingual work documents source-associated signals across 16 datasets and
nine languages~\citep{wang-etal-2026-human}. An EFL essay study finds that
rating-scale choice affects scores and agreement and that raters account for
the largest share of variability~\citep{barkaoui-2007-rating}. Human-evaluation studies identify
experimental design as a source of variation~\citep{clark-etal-2021-thats}, and
evaluation frameworks draw attention to cognitive biases
~\citep{elangovan-etal-2024-considers}. Evaluations of AI creative writing also
vary with reader priorities~\citep{marco-etal-2025-reader}. Human ratings are
products of a particular reader group and protocol, not an unqualified ground
truth.

\paragraph{Educational writing and source judgments.}
In a large study of English argumentative essays, teachers rated ChatGPT essays
above essays by non-native students, and computational analyses found
source-associated linguistic differences~\citep{herbold-etal-2023-large}.
Teachers and students nevertheless show limited source-detection
accuracy~\citep{waltzer-etal-2023-testing}, while actual and assumed source have
heterogeneous effects on analytic and holistic quality
ratings~\citep{fleckenstein-etal-2024-teachers}. Register-based analysis finds
that ChatGPT essays only partly reproduce the situational and linguistic
characteristics of undergraduate student writing
~\citep{goulart-etal-2024-ai}. Sociodemographic persona prompts similarly
produce measurable style differences~\citep{malik-etal-2024-empirical}, whereas student simulation
explicitly separates cognitive ability from writing
proficiency~\citep{chen-etal-2026-cpt}. A high generic writing score alone does
not show that a model writes at its assigned school level.

\paragraph{Korean writing and the KatFish corpus.}
Using Korean high-school essays, prior work compared student and GPT-4 writing
across morphological, syntactic, and sociolinguistic
measures~\citep{park-etal-2024-korean}. KatFishNet provides the closest
context for the present corpus. It combines multi-model linguistic profiling
with a separate three-reader quality audit of 75 texts. Its 30-essay component
uses seven three-point metrics and reports both higher LLM ratings and language
that often appeared more mature than the intended educational
level~\citep{park-etal-2025-katfishnet}. Under the source-informed scoring
protocol, the 16-criterion, 100-point total is itself the object of analysis.
We examine every criterion contribution, near-ceiling ratings, agreement within the
three-reader panel, source orderings by reader, and school-level score
profiles. This criterion-level account retains information about scoring
rules, linguistic values, and readers that the composite total does not
report.

\section{Study}

\subsection{Texts and Readers}

This paper presents a secondary, post hoc analysis; the research questions and
analyses were not preregistered. The study comprised 70 Korean argumentative
essays from KatFish: student essays and LLM essays attributed to GPT-4o, Solar,
Qwen2, and Llama3.1. The LLMs were prompted to write at an elementary-,
middle-, or high-school level. The three phases used disjoint essay sets.

All quantitative results use the 30 essays scored in Phase~2. Each of the five
sources contributes six essays, with two essays in every source--level
combination. Three undergraduates majoring in Korean language and literature
scored all 30 essays, producing $30\times3=90$ reader--essay evaluations. We
refer to them as readers throughout. The comparisons are descriptive of this
fixed sample.
Appendix~\ref{app:level-summary} reports the pooled and model-specific
school-level results together with the pairwise comparisons.

\subsection{Protocol and Rubric}
\label{sec:protocol-rubric}

Phase~1 asked readers to judge whether each essay was written by a student or
an LLM without using a writing rubric. After the source labels were disclosed,
the readers discussed recurring features and combined their observations with
the National Institute of Korean Language argumentative-writing diagnostic
framework~\citep{nikl-2025-argumentative}. According to the annotation
protocol, Phase~2 readers were not shown the gold source label of the essay
under evaluation. They knew the source categories and had seen the Phase~1
labels while helping construct the rubric. During Phase~2, they also recorded
an LLM, student, or uncertain source judgment for every essay. We refer to
Phase~2 as a \emph{source-informed scoring protocol}: the readers helped
construct the rubric after source disclosure and recorded a source judgment
while scoring every essay. The results characterize this protocol on the
30-essay sample and do not isolate textual quality from the source judgments
made during scoring.

The resulting study-specific rubric contains 16 scored criteria totaling 100
points: Content (40), Organization (10), and Expression (50), including two
creativity criteria. Phase~2 applied the same rubric at all three school levels. The NIKL
framework totals 45 points, or 40 when its first criterion is omitted, and was
developed through scoring and review of 12,000 writing
samples~\citep{nikl-2025-argumentative}. The present rubric changes its
criteria, anchors, and weights. Expression receives 50 of 100 points, compared
with 10 of 45 points in the NIKL framework, and the rubric adds two creativity criteria
worth 20 points in total. The study-specific rubric is not a direct application
of the official framework and has not been validated or normed by school level.
In this paper, \emph{rubric conformity} refers only to performance under these
scoring rules within the source-informed scoring protocol.

Phase~3 used a new, elementary-only sample after a second source-disclosure
stage and is excluded
from the school-level-balanced Phase~2 analysis. Appendix~\ref{app:protocol}
describes the three phases, and Appendix~\ref{app:rubric} provides the complete
translated rubric.

\paragraph{Analysis.}
Means, SDs, criterion contrasts, school-level summaries, pairwise comparisons,
weighting sensitivities, and joint-deletion checks use essay-level panel
means, each averaging the three readers. Creativity positive-score counts use
reader--essay scores, and ICCs use the full 30-essay $\times$ 3-reader matrix.
We report each source mean and the sample SD across its six essay-level panel
means. A pooled LLM mean averages all 24 LLM essays; the balanced design gives
every model and school level equal weight. School-level gaps subtract the
two-essay student mean from the eight-essay pooled LLM mean. Appendix~\ref{app:level-summary} defines the within-level comparisons.
For criterion $j$, the additive point contribution is
$\Delta_j=\bar{x}_{\mathrm{LLM},j}-\bar{x}_{\mathrm{Student},j}$; the 16
contributions sum exactly to the total-score gap. A negative $\Delta_j$ reduces
the pooled LLM advantage; \emph{student-leading} means that the student source
mean is highest or tied highest among the five sources. We summarize absolute
agreement for the fixed three-reader panel with two-way mixed ICC(A,1) and
ICC(A,3)~\citep{mcgraw-wong-1996-icc,mcgraw-wong-1996-correction}. The analyses
are descriptive of the fixed sample and scoring protocol. Derived quantities
use unrounded values. Displayed score means use decimal half-up rounding to two
places, percentages use one decimal where reported, and ICCs use three
decimals; independently rounded components need not sum exactly to the
displayed total.

\section{Results}

\subsection{Total Scores Under the Protocol}

Table~\ref{tab:scores} shows the same pattern for all four models. Every model
mean exceeds the student mean on the total and all three rubric dimensions.
The pooled LLM total is 71.79, compared with 53.44 for the student essays, a
difference of 18.35 points. The direction is also the same for every
reader~(Appendix~\ref{app:reader-totals}) and under the alternative weighting
and anchor treatments~(Appendix~\ref{app:sensitivity}).
\begin{table}[t]
\centering
\small
\setlength{\tabcolsep}{2.5pt}
\begin{tabular}{lrrrr}
\toprule
Source & Total /100 & C /40 & O /10 & E /50 \\
\midrule
Student essays & $53.44\pm11.65$ & 20.17 & 5.44 & 27.83 \\
GPT-4o         & $74.11\pm 5.39$ & 26.56 & 9.67 & 37.89 \\
Solar          & $67.67\pm 7.64$ & 24.56 & 8.67 & 34.44 \\
Qwen2          & $75.94\pm 6.48$ & 29.17 & 9.89 & 36.89 \\
Llama3.1       & $69.44\pm 7.93$ & 24.56 & 8.11 & 36.78 \\
\bottomrule
\end{tabular}
\caption{Mean total and dimension scores by essay source under the Phase~2
source-informed scoring protocol. Each LLM source mean exceeds the student
mean in all four columns. Totals are mean $\pm$ sample SD across six essays
after averaging the three reader scores for each essay. C, O, and E denote
Content, Organization, and Expression; columns are rounded independently to
two decimals.}
\label{tab:scores}
\end{table}
Because each reader shows the same direction, averaging does not create the
pooled student--LLM difference. Its meaning remains limited to performance
under this rubric and protocol; the rubric assigns half of its points to
Expression.
Total-score absolute agreement is ICC(A,1) $.605$ and ICC(A,3) $.821$;
Appendix~\ref{app:agreement} reports the intervals.

The school-level totals tell a different story from the overall ranking. The
student mean rises by 21.50 points from elementary to high school, while the
non-monotonic pooled LLM profile narrows the gap from 24.75 to 8.63 points. All
12 model-by-level means exceed the corresponding student mean, and an LLM essay
scores above a student essay in 47 of the 48 within-level comparisons. The
pooled LLM profile does not reproduce the rising student profile. With two
student essays per level, the results do not show that models match their
assigned school levels or follow a developmental pattern. Appendix~\ref{app:level-summary} gives the detailed results by essay
source.

\subsection{Near-Ceiling Formal Ratings and Reader Disagreement}

Twelve of the 16 criteria increase the pooled LLM--student gap, while four
reduce it: content creativity~($-3.06$), natural Korean
phrasing~($-1.18$), expression creativity~($-0.75$), and lexical
non-redundancy~($-0.06$). The five largest positive contributions are
orthographic norms~($+4.72$), 
genre-appropriate register~($+3.47$), paragraph
completeness~($+2.47$), sentence construction~($+2.15$), and evidence
sufficiency~($+2.06$).

Two formal criteria supply nearly half of the net point difference. Using
unrounded means, orthographic norms and genre-appropriate register contribute
44.7\% of the signed net gap: all 72 LLM ratings reach the orthographic maximum,
and 70 of 72 reach the register maximum. The orthographic criterion uses
explicit thresholds, while the register criterion has two printed anchors.
Creativity criteria and natural Korean phrasing also use coarse categories,
whereas agreement is substantially
lower. This contrast concerns the composition of the total; the study does not
record why a reader assigned an individual value.

For three of the four negative contributions, the student source mean is
highest or tied highest. Students lead on natural Korean phrasing~(3.89; model
range 2.22--3.06) and content creativity~(4.44; model range 0.56--2.78), and
tie Qwen2 on expression creativity~(1.39). Lexical non-redundancy is the
exception: its pooled contribution is $-0.06$, although Qwen2 has the highest
source mean. The low ICCs~(.091--.185 for ICC(A,1); .231--.405 for ICC(A,3))
and source-mean rankings that differ by reader do not support a shared
criterion-level ranking within this panel. Appendix
~\ref{app:reader-heterogeneity} reports the underlying means and counts.

\subsection{Interpreting the Total Score}

The total and its components tell different stories. ICC(A,3) is $.821$ for
the total, compared with $.231$ for natural Korean phrasing and $.405$ for
content creativity. The school-level totals likewise do not show that models
match their assigned levels, given two student essays per level and the
non-monotonic pooled LLM profile. The two near-ceiling formal criteria
contribute 44.7\% of the 18.35-point gap under this protocol. Claims about
creativity, natural Korean phrasing, and assigned school levels
require evidence beyond the total score. The decomposition reports point contributions in this sample,
not explained variance or general linguistic importance.

\section{Conclusion}

Comparing human and LLM essays requires more than reporting which source
receives the higher score. Criterion-level auditing shows what a rubric
rewards, where its scales restrict judgment, and where readers disagree. This
view helps researchers distinguish polished formal performance from
creativity, natural Korean phrasing, and other valued qualities. It also
supports comparisons across languages, genres, and reader groups. Transparent
rubrics keep human values visible.

\section*{Limitations}

This study examines Korean argumentative essays, a genre in which paragraph
structure, evidence, and formal register are explicit evaluation targets.
Holding language, genre, scoring rubric, and reader panel constant enables a
focused analysis. The study-specific rubric is neither validated nor normed by
school level and is examined as an evaluation object rather than an
educational standard. The resulting profile does not directly extend to
narrative, expository, reflective, creative, or informal writing. Comparative
studies should develop genre-appropriate rubrics and test which patterns persist
across forms, languages, and educational contexts.

The quantitative analysis uses a fixed set of 30 essays: six student essays
and 24 LLM essays, balanced across the five sources and three school
levels. This structure supports the present criterion-level case study because
every source and level was evaluated under the same scoring protocol. It does
not support population estimates or developmental conclusions, particularly
with two student essays at each level. Preregistered follow-up studies should
expand student coverage within each level and evaluate larger, independently
sampled sets before testing population-level or developmental hypotheses.

The same three undergraduates majoring in Korean language and literature
evaluated every essay. Their disciplinary background is relevant to Korean
form, register, and expression. Because all three readers scored every essay,
criterion comparisons are not confounded by changes in panel composition. A
three-person panel nevertheless represents one reader group with a shared
academic background rather than the range of perspectives involved in
educational writing. Follow-up panels should include Korean-language teachers, assessment
specialists, professional writers, students, and general readers, then report
how agreement, criterion priorities, and composite conclusions vary across
reader groups.

\section*{Ethical Considerations}

Rubric scores and source judgments can be misused to accuse individual
students of using AI. The evidence reported here concerns aggregate patterns in
a fixed research sample and does not establish the authorship of any individual
essay. These scores therefore provide no basis for academic-integrity
allegations or disciplinary decisions. We reject any use of this rubric, its
total score, or the source-associated tendencies discussed here as an
accusation mechanism.

\bibliography{custom}

\appendix

\section{Corpus Composition and Staged Protocol}
\label{app:protocol}

The reader study comprised three phases with disjoint essay sets.
Phase~1 collected source judgments without a writing rubric. After source
labels were disclosed, the readers discussed recurring features and
constructed the study-specific rubric. Phase~2 used the rubric to score a new
sample balanced by source and school level; all quantitative results in this
paper come from this phase. Phase~3 used another new sample after a second
source-disclosure stage and included only elementary-school texts. Because its
sample composition and prior reader exposure differ from Phase~2, no Phase~3
score enters the reported analysis.

All essays were drawn from the previously generated KatFish corpus; no new LLM
essays were generated for this analysis. LLM essays in the two scoring phases are attributed to GPT-4o,
Solar, Qwen2, and Llama3.1. KatFishNet documents the Korean generation template
in its Appendix~C, Table~6~\citep{park-etal-2025-katfishnet}. In paraphrase,
the template asks a model to assume a specified school level, write only a
Korean argumentative essay for the supplied topic and prompt, and return only
the essay.

Table~\ref{tab:composition} summarizes the staged corpus. Essay identifiers do
not repeat across phases. All three readers evaluated every essay, so the
Evaluations column equals the number of essays multiplied by three.

\begin{table}[tbp]
\centering
\setlength{\tabcolsep}{3pt}
\begin{tabular}{lrrrr}
\toprule
Phase & Student & LLM & Total & Evaluations \\
\midrule
1 & 6 & 24 & 30 & 90 \\
2 & 6 & 24 & 30 & 90 \\
3 & 2 &  8 & 10 & 30 \\
\midrule
All & 14 & 56 & 70 & 210 \\
\bottomrule
\end{tabular}
\caption{Essay and reader--essay evaluation counts across the three protocol
phases. All three readers evaluated every essay, so Evaluations is three times
Total.}
\label{tab:composition}
\end{table}

The 210 judgments come from the same three-reader panel and are not 210
independent observations. Phase~2 contributes 90 reader--essay evaluations of
16 scored criteria organized into 11 groups: thesis clarity, evidence,
counterargument, content creativity, paragraph-level organization, sentence
cohesion, lexical usage, sentence construction, orthographic and
standard-language norms, writing conventions, and expression creativity.

\section{Operational Rubric and Anchor Structure}
\label{app:rubric}

Tables~\ref{tab:rubric-content-organization} and
\ref{tab:rubric-expression} reproduce the study-specific 16-criterion, 100-point
rubric in English translation. The rubric assigns 40 points to Content, 10 to
Organization, and 50 to Expression. It was adapted from the National Institute
of Korean Language framework~\citep{nikl-2025-argumentative} under Korea Open
Government License Type~1. The panel changed the criteria, anchors, and weights
and added two creativity criteria. These changes distinguish the study-specific
rubric from a direct application of the NIKL framework.

Score interpretation depends on substantial differences among the printed
response scales. Several criteria list three ordered anchors, evidence
sufficiency and evidence depth provide descriptions at the endpoints, content
creativity lists 0 and 10, and expression creativity lists 0, 5, and 10. The total thus
combines criteria that differ in both judgment type and score granularity.

\begin{table*}[t]
\centering
\footnotesize
\setlength{\tabcolsep}{4pt}
\renewcommand{\arraystretch}{1.08}
\begin{tabularx}{\textwidth}{@{}>{\raggedright\arraybackslash}p{0.20\textwidth}>{\raggedright\arraybackslash}p{0.22\textwidth}Y@{}}
\toprule
Rubric group & Criterion & Printed score anchors \\
\midrule
\multicolumn{3}{@{}l}{\textbf{Content (40 points)}} \\
Thesis clarity (10) & Alignment with the prompt &
5: thesis is well aligned; 3: parts are not aligned; 1: essay largely deviates from the prompt. \\
Thesis clarity (10) & Thesis consistency &
5: thesis is consistent throughout; 3: parts are inconsistent; 1: thesis is inconsistent. \\
Evidence appropriateness and sufficiency (15) & Appropriateness of evidence &
5: evidence is appropriate; 3: at least one piece is inappropriate; 1: two or more pieces are inappropriate. \\
Evidence appropriateness and sufficiency (15) & Sufficiency of evidence &
5: evidence is sufficient; 1: evidence is present yet insufficient. \\
Evidence appropriateness and sufficiency (15) & Depth of evidence &
5: evidence is elaborated with depth; 1: evidence is present without depth. \\
Counterargument consideration (5) & Introducing and evaluating counterarguments &
5: a counterargument is introduced and addressed in depth; 3: one is introduced; 1: none is introduced. \\
Creativity (10) & Creative content & 10: yes; 0: no. \\
\midrule
\multicolumn{3}{@{}l}{\textbf{Organization (10 points)}} \\
Overall paragraph organization (5) & Within-paragraph structure &
5: intro--body--conclusion structure; 3: somewhat incomplete; 1: lacks completeness. \\
Inter-sentence cohesion (5) & Logical sentence connections &
5: tightly connected; 3: some connections are illogical; 1: cohesion is weak for most links. \\
\bottomrule
\end{tabularx}
\caption{Study-specific 100-point rubric, Part I: Content and Organization.
Parentheses show the maximum score for each rubric group; the highest printed
anchor shows the maximum for each criterion. English translation by the
authors.}
\label{tab:rubric-content-organization}
\end{table*}

Content creativity contributes one quarter of the 40-point Content dimension
score, while its printed scale lists 0 and 10. Evidence sufficiency and evidence
depth provide verbal descriptions only for the endpoints of their five-point
spans. Organization is narrower: its two five-point criteria assess paragraph
completeness and sentence cohesion. Equal point differences across these groups
therefore do not represent equally granular judgments.

The 50-point Expression dimension combines explicit error-rate thresholds for
sentence construction and orthographic norms with categorical judgments of
genre-appropriate register, natural Korean phrasing, and expression
creativity. Interpreting the total therefore requires attention to both criterion
weights and response-scale structure.

\begin{table*}[t]
\centering
\footnotesize
\setlength{\tabcolsep}{4pt}
\renewcommand{\arraystretch}{1.08}
\begin{tabularx}{\textwidth}{@{}>{\raggedright\arraybackslash}p{0.20\textwidth}>{\raggedright\arraybackslash}p{0.22\textwidth}Y@{}}
\toprule
Rubric group & Criterion & Printed score anchors \\
\midrule
\multicolumn{3}{@{}l}{\textbf{Expression (50 points)}} \\
Lexical use (10) & Lexical accuracy &
5: inaccurate use below 1\% of eojeols; 3: 1\% to below 3\%; 1: at least 3\%. \\
Lexical use (10) & Unnecessary lexical repetition &
5: appropriate; 0: inappropriate. \\
Sentence construction (10) & Length, required constituents, grammatical errors, and ambiguity &
10: awkward or incorrect portions below 10\% of sentences; 7: 10--20\%; 4: 20--30\%; 0: above 30\%. \\
Korean orthography and standard-language norms (10) & Spelling, standard-language usage, and related conventions &
10: errors below 1\% of total word count; 5: 1\% to below 3\%; 0: at least 3\%. \\
Writing conventions (10) & Genre-appropriate register, including colloquial expressions &
5: appropriate; 0: inappropriate. \\
Writing conventions (10) & Natural Korean phrasing, including translationese &
5: appropriate; 0: inappropriate. \\
Creativity (10) & Creative wording or style &
10: at least two instances; 5: one instance; 0: no instances. \\
\bottomrule
\end{tabularx}
\caption{Study-specific 100-point rubric, Part II: Expression. Parentheses show
the maximum score for each rubric group; the highest printed anchor shows the
maximum for each criterion. \emph{Eojeol} denotes a Korean spacing unit.
English translation by the authors.}
\label{tab:rubric-expression}
\end{table*}

The orthographic criterion gives 10 points whenever counted errors remain below
1\% of the word total. Every LLM rating reaches this maximum.
Genre-appropriate register prints 0 and 5, and 70 of 72 LLM ratings receive 5.
Natural Korean phrasing uses the same coarse printed scale, yet its direction
differs across readers. Expression creativity supplies a third scale shape and
produces sparse positive scores. The Expression total therefore combines
ratings produced by markedly different scale structures.

Among the $30\times3\times16=1{,}440$ Phase~2 criterion scores, 47 lie
numerically between the minimum and maximum of a criterion without matching a printed anchor. The
printed rubric does not state how such values should be interpreted. The
primary analysis retains the scores as given rather than retrospectively
recoding reader judgments; this choice does not treat the 47 values as
validated scale points. Lower- and upper-anchor mappings test numerical
sensitivity, and neither is asserted to be correct. The translated
rubric tables above provide the score ranges and printed anchors needed to
interpret the decomposition and sensitivity analyses.

\section{School-Level Results}
\label{app:level-summary}

Table~\ref{tab:level-detail} reports both the pooled and model-specific
profiles. Each level
contains two student essays and two essays from each model, producing
eight pooled LLM essays and 16 within-level LLM--student essay comparisons.

\begin{table*}[t]
\centering
\setlength{\tabcolsep}{4pt}
\begin{tabular}{lrrrrrrrr}
\toprule
Level & Student & GPT-4o & Solar & Qwen2 & Llama3.1 & Pooled LLM & Gap & Wins \\
\midrule
Elementary & 42.50 & 69.33 & 61.17 & 77.00 & 61.50 & 67.25 & 24.75 & 16/16 \\
Middle     & 53.83 & 77.67 & 72.67 & 79.83 & 71.83 & 75.50 & 21.67 & 16/16 \\
High       & 64.00 & 75.33 & 69.17 & 71.00 & 75.00 & 72.63 &  8.63 & 15/16 \\
\bottomrule
\end{tabular}
\caption{Mean Phase~2 total scores by school level and essay source. Gap is the
pooled LLM mean minus the student mean; Wins counts the 16 within-level
LLM--student essay comparisons in which the LLM essay receives the higher
total. Every LLM source mean exceeds the student mean at each level, while the
pooled gap narrows from elementary to high school.}
\label{tab:level-detail}
\end{table*}

All four model means exceed the student mean at every school level, yielding
12 of 12 source--level comparisons in the same direction. Across the 48
within-level comparisons, an LLM essay receives the higher score in 16
elementary, 16 middle-school, and 15 high-school pairs. These comparisons reuse
essays and are not independent trials.

The level profiles differ in shape. The two-essay student means rise from
42.50 to 53.83 and 64.00, a 21.50-point difference between the elementary and
high-school groups. The pooled LLM means are 67.25, 75.50, and 72.63,
peaking in the middle-school group. GPT-4o, Solar, and Qwen2 also peak at the
middle-school level, whereas Llama3.1 reaches its largest mean at the
high-school level. The pooled gap consequently narrows from 24.75 to 8.63
points.

These three school-level means do not establish a developmental trajectory.
Each student group contains two essays, and the same rubric was applied at
every school level. The results distinguish high rubric scores from evidence
that models match their assigned school levels.

\section{Complete Criterion Decomposition}
\label{app:criterion-means}

Table~\ref{tab:criterion-means} reports all 16 criteria used to form the
Content, Organization, and Expression totals in Table~\ref{tab:scores}. Each
source entry averages six essay-level panel means, and each panel mean averages
the same three reader scores. The pooled LLM mean gives equal weight to the
four balanced LLM source groups. For criterion \(j\),
\(\Delta_j=\bar{x}_{\mathrm{LLM},j}-\bar{x}_{\mathrm{Student},j}\), measured in
the rubric points available for that criterion.

\begin{table*}[t]
\centering
\setlength{\tabcolsep}{2.0pt}
\begin{tabular}{lrrrrrrr}
\toprule
Criterion (maximum) & Student & GPT-4o & Solar & Qwen2 & Llama3.1 & Pooled LLM & $\Delta_j$ \\
\midrule
Topic alignment (5) & 3.67 & 4.89 & \textbf{5.00} & \textbf{5.00} & \textbf{5.00} & 4.97 & $+1.31$ \\
Stance consistency (5) & 3.56 & 4.67 & 4.78 & \textbf{5.00} & 4.44 & 4.72 & $+1.17$ \\
Evidence relevance (5) & 3.44 & 4.78 & \textbf{5.00} & \textbf{5.00} & 4.56 & 4.83 & $+1.39$ \\
Evidence sufficiency (5) & 1.89 & \textbf{4.61} & 3.89 & 4.22 & 3.06 & 3.94 & $+2.06$ \\
Evidence depth (5) & 1.61 & 3.78 & 2.78 & \textbf{4.17} & 2.83 & 3.39 & $+1.78$ \\
Counterargument (5) & 1.56 & \textbf{3.28} & 2.56 & 3.00 & 3.00 & 2.96 & $+1.40$ \\
Content creativity (10) & \textbf{4.44} & 0.56 & 0.56 & 2.78 & 1.67 & 1.39 & $-3.06$ \\
\midrule
Paragraph completeness (5) & 2.11 & \textbf{4.89} & 4.56 & \textbf{4.89} & 4.00 & 4.58 & $+2.47$ \\
Sentence cohesion (5) & 3.33 & 4.78 & 4.11 & \textbf{5.00} & 4.11 & 4.50 & $+1.17$ \\
\midrule
Lexical appropriateness (5) & 4.67 & \textbf{5.00} & \textbf{5.00} & 4.89 & \textbf{5.00} & 4.97 & $+0.31$ \\
Lexical non-redundancy (5) & 4.22 & 4.44 & 3.33 & \textbf{4.72} & 4.17 & 4.17 & $-0.06$ \\
Sentence construction (10) & 7.00 & \textbf{9.83} & 8.61 & 8.67 & 9.50 & 9.15 & $+2.15$ \\
Orthographic norms (10) & 5.28 & \textbf{10.00} & \textbf{10.00} & \textbf{10.00} & \textbf{10.00} & 10.00 & $+4.72$ \\
Genre-appropriate register (5) & 1.39 & \textbf{5.00} & \textbf{5.00} & 4.72 & 4.72 & 4.86 & $+3.47$ \\
Natural Korean phrasing (5) & \textbf{3.89} & 3.06 & 2.22 & 2.50 & 3.06 & 2.71 & $-1.18$ \\
Expression creativity (10) & \textbf{1.39} & 0.56 & 0.28 & \textbf{1.39} & 0.33 & 0.64 & $-0.75$ \\
\bottomrule
\end{tabular}
\caption{Mean Phase~2 criterion scores by essay source. For criterion \(j\),
$\Delta_j$ is the pooled LLM mean minus the student mean; 12 criteria increase
the total-score gap and four reduce it. Bold marks the highest mean among the
five individual essay sources in each row, including all ties. Contributions
use unrounded means; displayed scores are rounded to two decimals.}
\label{tab:criterion-means}
\end{table*}

Using unrounded means, the 12 positive contributions sum to 23.39 points and
the four negative contributions offset 5.04 points, leaving the reported
18.35-point net gap after rounding. A negative pooled contribution does not
imply that the student mean is largest among every individual source. Lexical
non-redundancy illustrates the distinction: its pooled contribution is
\(-0.06\), whereas Qwen2 has the largest source mean.

\paragraph{Argumentation and organization.}
The pooled LLM mean exceeds the student mean on all six non-creativity Content
criteria.
Evidence sufficiency contributes \(+2.06\), evidence depth \(+1.78\), and
counterargument handling \(+1.40\). Content creativity moves in the opposite
direction by \(3.06\) points and is the largest negative contribution.
Organization adds \(+2.47\) for paragraph completeness and \(+1.17\) for
sentence cohesion. At the dimension level, positive contributions from
conventional argument execution coexist with a 3.06-point negative
contribution from content creativity.

\paragraph{Expression.}
The largest positive contribution is orthographic norms at \(4.72\) points;
every one of the 72 LLM ratings receives the maximum. Genre-appropriate
register adds \(3.47\) points, with 70 of 72 LLM ratings at its
maximum. Sentence construction adds another \(2.15\). Natural Korean phrasing,
expression creativity, and lexical non-redundancy offset \(1.18\), \(0.75\),
and \(0.06\) points, respectively. The Expression total combines near-ceiling
formal ratings with lower pooled LLM means for natural Korean phrasing,
expression creativity, and lexical non-redundancy.

The two displayed formal contributions sum to 8.19 points. Using unrounded
criterion means, orthographic norms and genre-appropriate register account for
44.7\% of the signed net gap. The denominator is the unrounded net gap after
negative contributions have reduced the total. This percentage is neither
explained variance nor a measure of the broader linguistic importance of the
two criteria. The decomposition shows how the rubric and its weights produce
the observed score, not why a reader selected a particular value.

\section{Inter-reader Agreement}
\label{app:agreement}

Table~\ref{tab:agreement} reports absolute-agreement, two-way mixed
intraclass correlations because the same three readers rated every Phase~2
essay and those readers form the panel of
interest~\citep{mcgraw-wong-1996-icc,mcgraw-wong-1996-correction}. ICC(A,1) describes
the score from one observed reader. ICC(A,3) describes the mean of all three
readers. The 95\% intervals use the corresponding \(F\)-distribution
approximation.

\begin{table*}[tbp]
\centering
\setlength{\tabcolsep}{2.2pt}
\begin{tabular}{lcc}
\toprule
Score & ICC(A,1) [95\% CI] & ICC(A,3) [95\% CI] \\
\midrule
Total & .605 [.408, .768] & .821 [.674, .908] \\
Content & .355 [.136, .581] & .623 [.320, .806] \\
Organization & .621 [.428, .779] & .831 [.692, .913] \\
Expression & .492 [.276, .689] & .744 [.534, .869] \\
Natural Korean phrasing & .091 [$-$.057, .299] & .231 [$-$.194, .562] \\
Content creativity & .185 [$-$.023, .431] & .405 [$-$.072, .694] \\
Expression creativity & .158 [$-$.048, .406] & .361 [$-$.158, .672] \\
\bottomrule
\end{tabular}
\caption{Absolute agreement among the same three readers for the 30 Phase~2
essays. ICC(A,1) summarizes a single reader score, and ICC(A,3) summarizes the
mean of the three reader scores. Agreement is higher for the total than for
natural Korean phrasing and the two creativity criteria.}
\label{tab:agreement}
\end{table*}

Among the aggregate dimensions, agreement is strongest for Organization:
ICC(A,1) is .621 and ICC(A,3) is
.831. The total is similar for the three-reader mean at .821, whereas Content
is lower at .623. Agreement is higher for the three-reader mean than for a
single reader, whereas it remains low for natural Korean phrasing, content creativity,
and expression creativity. Their ICC(A,3) values are .231, .405, and .361,
and all three intervals include zero.

These estimates describe the fixed reader panel on the 30-essay sample.
Between-source and between-level dispersion contributes to the coefficients,
and sparse or ceiling-heavy scores increase uncertainty. The
intervals are model-based summaries rather than design-based population
intervals. Following the published correction, ICC(A,3) and its interval
endpoints are obtained from the corresponding ICC(A,1) value \(\rho\) through
\(3\rho/(1+2\rho)\).

\subsection{Reader-Level Total Scores}
\label{app:reader-totals}

The student--LLM direction in Table~\ref{tab:scores} is present for every
reader, so averaging across the panel does not create the separation.
Table~\ref{tab:reader-totals} gives the reader-level scores needed to assess
its magnitude. Each of the 12 reader--model comparisons is positive, with
differences ranging from 9.33 points for A2 on Solar to 31.33 points for A1 on
Qwen2.

\begin{table*}[t]
\centering
\setlength{\tabcolsep}{4pt}
\begin{tabular}{lccccc}
\toprule
Reader & Student & GPT-4o & Solar & Qwen2 & Llama3.1 \\
\midrule
A1 & $50.67\pm19.61$ & $76.50\pm4.09$ & $74.00\pm4.05$ & $82.00\pm3.90$ & $70.00\pm12.41$ \\
A2 & $55.33\pm15.97$ & $73.67\pm4.27$ & $64.67\pm9.35$ & $72.50\pm10.56$ & $69.67\pm8.94$ \\
A3 & $54.33\pm6.28$ & $72.17\pm12.45$ & $64.33\pm12.08$ & $73.33\pm8.33$ & $68.67\pm7.63$ \\
\bottomrule
\end{tabular}
\caption{Phase~2 total scores by reader and essay source (mean $\pm$ sample SD
across six essays per source). For all three readers, every LLM source mean
exceeds the student mean, whereas LLM rankings and gap sizes vary by reader.}
\label{tab:reader-totals}
\end{table*}

A1 gives the highest LLM mean to Qwen2 and places Solar above Llama3.1. A2
places GPT-4o first and Llama3.1 above Solar, while A3 places Qwen2 first and
also ranks Llama3.1 above Solar. All three readers place student essays below
every LLM source in mean total score, whereas their LLM rankings and gap magnitudes
differ. The criterion-level results likewise show that a shared total-score
direction does not imply a shared criterion ordering.

\subsection{Criterion-Level Reader Heterogeneity}
\label{app:reader-heterogeneity}

Low agreement alone does not identify the direction of disagreement.
Table~\ref{tab:reader-heterogeneity} disaggregates the three exploratory
criteria selected after inspecting all 16 criteria. Each cell averages the six
essays assigned to a source by one reader. The table describes the three
observed readers rather than a reader population.

\begin{table*}[t]
\centering
\begin{tabular}{llrrrrr}
\toprule
Criterion & Reader & Student & GPT-4o & Solar & Qwen2 & Llama3.1 \\
\midrule
Natural Korean phrasing & A1 & 2.50 & 4.17 & 3.33 & 4.17 & 4.17 \\
                        & A2 & 4.17 & 0.00 & 0.83 & 0.83 & 0.83 \\
                        & A3 & 5.00 & 5.00 & 2.50 & 2.50 & 4.17 \\
\midrule
Content creativity & A1 & 1.67 & 0.00 & 0.00 & 3.33 & 0.00 \\
                   & A2 & 6.67 & 1.67 & 0.00 & 1.67 & 3.33 \\
                   & A3 & 5.00 & 0.00 & 1.67 & 3.33 & 1.67 \\
\midrule
Expression creativity & A1 & 0.00 & 0.00 & 0.00 & 1.67 & 0.17 \\
                      & A2 & 2.50 & 0.83 & 0.00 & 2.50 & 0.00 \\
                      & A3 & 1.67 & 0.83 & 0.83 & 0.00 & 0.83 \\
\bottomrule
\end{tabular}
\caption{Phase~2 mean scores by reader and essay source for three exploratory
criteria. Each cell averages six essays. Rankings for natural Korean phrasing
differ across readers.}
\label{tab:reader-heterogeneity}
\end{table*}

Natural Korean phrasing exhibits a directional reversal. A1 assigns every
model a higher mean than the student essays. A2 assigns the student essays a
higher mean than every model. A3 ties the student and GPT-4o means and places
the student mean above Solar, Qwen2, and Llama3.1. The pooled student advantage
therefore does not represent a shared ranking rule.

Content creativity also varies in magnitude. A1 assigns student essays a mean
of 1.67 and Qwen2 a mean of 3.33, whereas A2 and A3 assign student essays the
largest means, 6.67 and 5.00. Expression-creativity scores are sparse across
every source and reader. Table~\ref{tab:creativity-counts} consequently reports
positive-rating and essay-level majority counts rather than relying only on
means.

\begin{table}[tbp]
\centering
\small
\setlength{\tabcolsep}{2pt}
\begin{tabular}{lrrrrrr}
\toprule
& \multicolumn{3}{c}{Content creativity} & \multicolumn{3}{c}{Expression creativity} \\
\cmidrule(lr){2-4}\cmidrule(lr){5-7}
Source & $+$/18 & Any/6 & Maj./6 & $+$/18 & Any/6 & Maj./6 \\
\midrule
Student  & 8 & 5 & 3 & 5 & 4 & 1 \\
GPT-4o   & 1 & 1 & 0 & 2 & 2 & 0 \\
Solar    & 1 & 1 & 0 & 1 & 1 & 0 \\
Qwen2    & 5 & 3 & 2 & 4 & 2 & 2 \\
Llama3.1 & 3 & 2 & 1 & 2 & 1 & 1 \\
\bottomrule
\end{tabular}
\caption{Counts of positive creativity scores. $+$/18 is the number of
reader--essay scores above zero, Any/6 is the number of essays with at least
one positive score, and Maj./6 is the number with positive scores from at
least two readers.}
\label{tab:creativity-counts}
\end{table}

For content creativity, student essays receive 8 of 18 positive ratings, and
three of six receive positive scores from at least two readers. Qwen2 is the
closest model source with 5 of 18 ratings and two majority-positive essays.
For expression creativity, student essays have one more positive rating than
Qwen2, whereas Qwen2 has two majority-positive essays compared with one
student essay. Pooled means, rating counts, and essay-majority counts therefore
yield different summaries. None provides reader-independent evidence that
creativity consistently favors one source.

\section{Weighting, Anchor, and Deletion Sensitivity}
\label{app:sensitivity}

Table~\ref{tab:weight-sensitivity} examines four post hoc alternatives to the
reported total. Let \(x_{ij}\) denote the score for essay \(i\) on criterion
\(j\), and let \(m_j\) denote the maximum for that criterion. Equal-criterion
weighting is
\(\frac{100}{16}\sum_{j=1}^{16}x_{ij}/m_j\). Equal-dimension weighting is
\(\frac{100}{3}(C_i/40+O_i/10+E_i/50)\), where \(C_i\), \(O_i\), and \(E_i\)
are the three reported dimension scores. The lower-anchor mapping replaces each
of the 47 between-anchor values with the adjacent lower printed anchor; the
upper-anchor mapping uses the adjacent upper printed anchor. These are
descriptive sensitivity checks rather than validated replacement scales.

\begin{table}[tbp]
\centering
\small
\setlength{\tabcolsep}{2.2pt}
\begin{tabular}{lrrrrr}
\toprule
Source & Reported & Eq. crit. & Eq. dims. & Lower & Upper \\
\midrule
Student  & 53.44 & 55.49 & 53.51 & 52.78 & 54.39 \\
GPT-4o   & 74.11 & 79.55 & 79.61 & 73.00 & 74.89 \\
Solar    & 67.67 & 72.43 & 72.31 & 66.33 & 69.22 \\
Qwen2    & 75.94 & 80.66 & 81.86 & 74.61 & 77.06 \\
Llama3.1 & 69.44 & 73.37 & 72.02 & 68.83 & 70.44 \\
\bottomrule
\end{tabular}
\caption{Mean source scores under the reported total and four post hoc
alternatives. Eq. crit. assigns equal weight to the 16 criteria, and Eq. dims.
assigns equal weight to Content, Organization, and Expression. Lower and Upper
map the 47 between-anchor values to the adjacent lower or upper printed anchor.
The student mean remains lowest under all four alternatives.}
\label{tab:weight-sensitivity}
\end{table}

The student mean remains lowest under every alternative. Using unrounded
values, equal-criterion weighting raises the source means by 2.04--5.44 points, and
equal-dimension weighting changes them by 0.06--5.92 points. The latter
weighting reverses the relative order of Solar and Llama3.1, so the table
supports persistence of the student--LLM separation rather than complete rank
invariance among LLMs. Mapping the 47 between-anchor values downward
changes source means by at most 1.33 points, and mapping them upward changes source means by
at most 1.56 points. The values in the table are rounded to two decimals after
each source mean is computed.

Table~\ref{tab:fixed-robustness} adds joint-deletion and direct-comparison
checks. For each model, we delete one of its six essays and one of the six
student essays, then subtract the remaining five-essay student mean from the
remaining five-essay LLM mean. The 36 possible deletion pairs produce 36
recomputed gaps for each LLM. The minimum column reports the smallest of these
gaps, and the Wins column counts the 36 direct LLM--student essay comparisons.

\begin{table}[tbp]
\centering
\setlength{\tabcolsep}{4pt}
\begin{tabular}{lrrr}
\toprule
Source & Gap & Min. deletion gap & Wins \\
\midrule
GPT-4o   & 20.67 & 16.80 & 35/36 \\
Solar    & 14.22 & 10.00 & 32/36 \\
Qwen2    & 22.50 & 17.27 & 36/36 \\
Llama3.1 & 16.00 & 11.80 & 33/36 \\
\bottomrule
\end{tabular}
\caption{Robustness of the LLM--student total-score gap to essay deletion. Gap
is the reported LLM mean minus the student mean; Min. deletion gap is the
smallest recomputed gap after deleting one LLM essay and one student essay;
Wins counts the 36 direct essay pairs in which the LLM essay receives the
higher score. All 144 deletion combinations retain a positive mean gap.}
\label{tab:fixed-robustness}
\end{table}

Tables~\ref{tab:weight-sensitivity} and~\ref{tab:fixed-robustness} perturb
different parts of the comparison. Reweighting and anchor mapping keep the
essay set fixed while changing score construction. Joint deletion keeps the
reported rubric fixed while changing the particular five-essay subsets being
compared. Across the weighting, anchor, and deletion checks, all four
student--LLM gaps remain positive within this sample.

All 144 deletion combinations retain a positive mean gap. Qwen2 shows the
largest minimum gap and 36 of 36 direct wins; Solar shows the smallest minimum
gap and 32 of 36 wins. The exact model hierarchy is less stable:
equal-dimension weighting switches Solar and Llama3.1, direct wins range from
32/36 to 36/36, and minimum deletion gaps range from 10.00 to 17.27 points.
The four student--LLM contrasts retain the same direction, while LLM ordering
and gap magnitude vary across checks.

These checks address within-sample stability, not validity. They test whether
the observed source means change direction under the specified weights, anchor
mappings, and essay deletions. They do not establish comprehensive coverage of
Korean writing quality, agreement for another reader panel, or replication in
a new sample. The criterion decomposition and ICC estimates remain necessary
for interpreting the consistent composite pattern.

\end{document}